\documentclass[10pt,twocolumn,letterpaper]{article}

\usepackage{cvpr}
\usepackage{times}
\usepackage{epsfig}
\usepackage{graphicx}
\usepackage{amsmath}
\usepackage{amssymb}
\usepackage{url}
\usepackage{caption}
\usepackage{framed}

\cvprfinalcopy 

\def\cvprPaperID{4} 

\begin{document}

\title{Intel\textsuperscript{\textregistered\ } RealSense\texttrademark\ Stereoscopic Depth Cameras}
\author{Leonid Keselman \qquad John Iselin Woodfill \qquad Anders Grunnet-Jepsen \qquad Achintya Bhowmik \\
Intel Corporation \\
{\tt\small \{leonid.m.keselman | j.i.woodfill | anders.grunnet-jepsen | achintya.k.bhowmik\}@intel.com}
}

\maketitle

\begin{abstract}
We present a comprehensive overview of the stereoscopic Intel RealSense RGBD imaging systems. We discuss these systems' mode-of-operation, functional behavior and include models of their expected performance, shortcomings, and limitations. We provide information about the systems' optical characteristics, their correlation algorithms, and how these properties can affect different applications, including 3D reconstruction and gesture recognition. Our discussion covers the Intel RealSense R200 and the Intel RealSense D400 (formally RS400). 
\end{abstract}

\section{Introduction}
Portable, consumer-grade RGBD systems gained popularity with the Microsoft Kinect. By including hardware-accelerated depth computation over a USB connection, it kicked off wide use of RGBD sensors in computer vision, human-computer interaction and robotics. In 2015, Intel announced a family of stereoscopic, highly portable, consumer, RGBD sensors that include subpixel disparity accuracy, assisted illumination, and function well outdoors. 

Previously documented features of the Intel RealSense R200 were limited to an in-depth discussion of the electrical, mechanical and thermal properties~\cite{intel:r200mouser}, or high-level usage information when utilizing the provided software development kit and designing end-user applications~\cite{intel:r200sdk}.  In contrast, this paper presents a technical overview of the imaging and computation systems in this line of products.

\begin{figure}
\centering
\begin{framed}
  \includegraphics[width=0.8\linewidth]{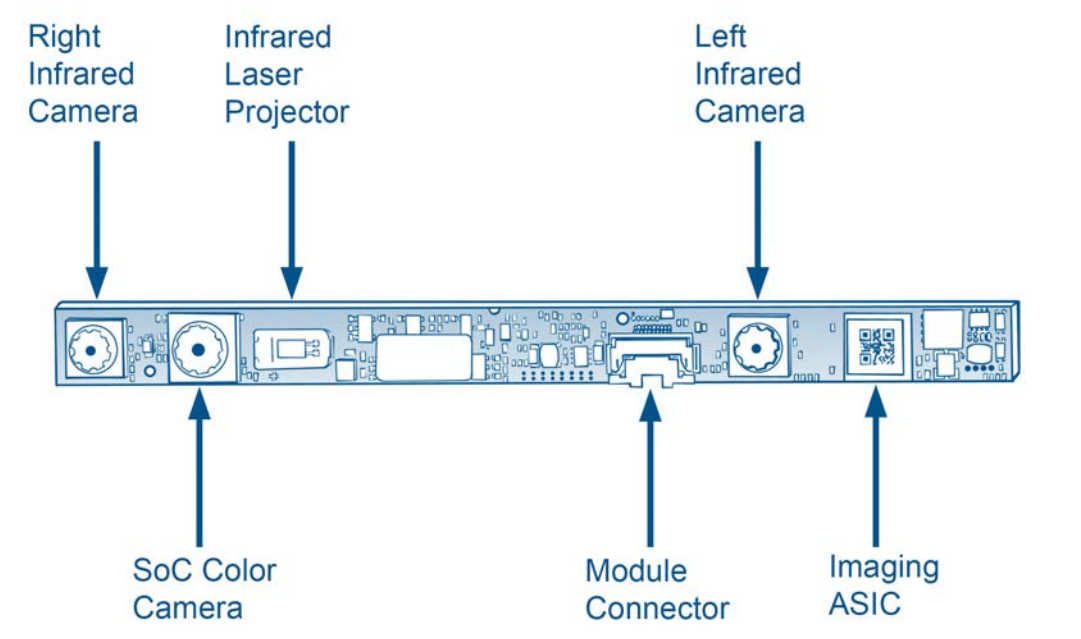} \\
    \vspace{5mm}
  \includegraphics[width=\linewidth]{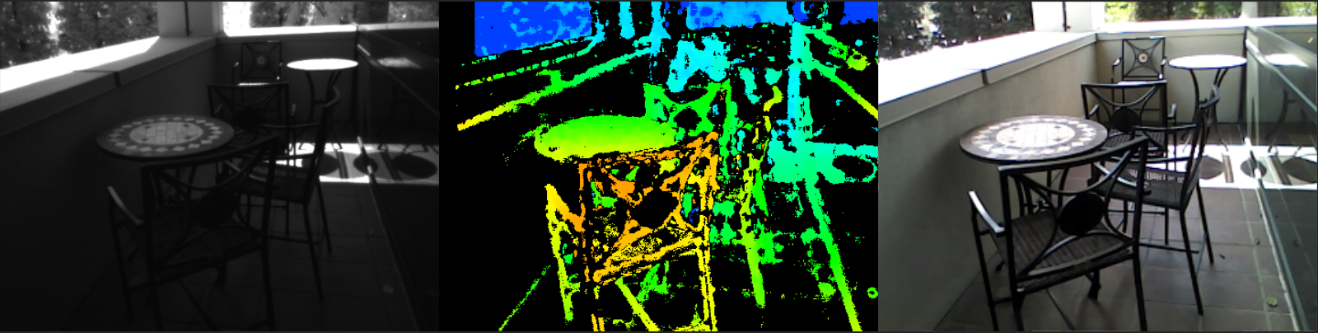}
  \includegraphics[width=\linewidth]{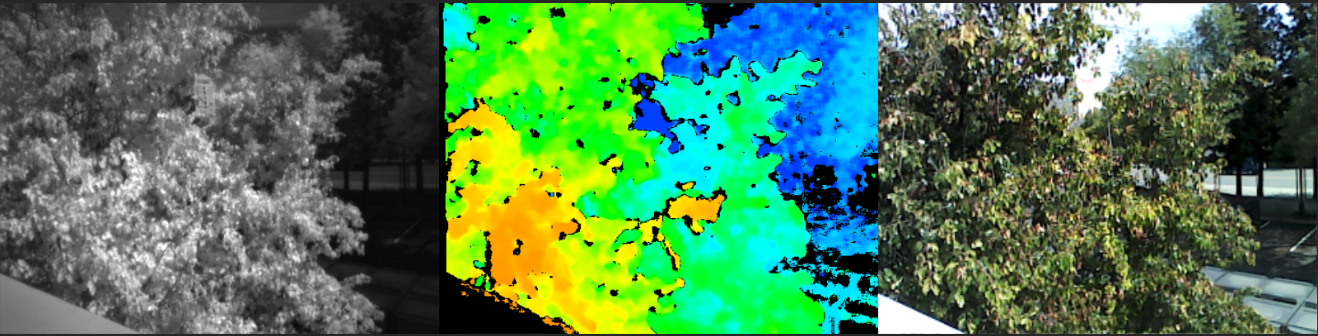}
\end{framed}
\caption{An overview of a RealSense R200 module, along with examples of the data captured in outdoor conditions. From left to right are the infrared, depth and color images.}
\label{fig:ds4}
\end{figure}

\section{Theory of Operation}
\subsection{Stereo Depth}
In general, the relationship between a disparity ($d$) and depth ($z$) can be parametrized as seen in equation \ref{eq:dispdepth}. Here focal length of the imaging sensor ($f$) is in pixels, while the baseline between the camera pair ($B$) is in the desired depth units (typically \textbf{m} or \textbf{mm}).

\begin{equation} \label{eq:dispdepth}
z = \frac{f \cdot B}{d}
\end{equation}
Additionally, we can take the derivative, $\frac{\partial z}{\partial d}$, and substitute in the relationship between $z$ and $d$ to generate an error relationship with distance.
\begin{equation} \label{eq:derr}
|\epsilon_{z}| = \frac{z^2}{f \cdot B} \cdot |\epsilon_{d}|
\end{equation}
Given that errors in disparity space are usually constant for a stereo system, stemming from imaging properties (such as SNR and MTF) and the quality of the matching algorithm, we can treat $|\epsilon_{d}|$ as a constant in passive systems ($\approx 0.1$ from measurements in Section~\ref{sec:perf}) \footnote{In active systems, the projector's $\frac{1}{z^2}$ falloff leads to lower SNR at small disparities (e.g. longer distances), and therefore $|\epsilon_{d}|$ is only approximately constant until imaging noise overwhelms projector intensity.}.

\subsection{Unstructured Light}
Classical stereoscopic depth systems struggle with resolving depth on texture-less surfaces. A plethora of techniques have been developed to solve this problem, from global optimization methods, to semi-global propagation techniques, to plane sweeping methods. However, these techniques all depend on some prior assumptions about data in order to generate correct depth candidates. In the Intel RGBD depth cameras, there is instead an active texture projector available on the module. This technique was used by classical stereo systems, going back to 1984, and  was originally called \textit{unstructured light}~\cite{nishihara1984practical,Nishihara:1984:PPR:889286}.

Such systems do not require a priori knowledge of the pattern's structure, as they are used simply to generate texture which makes image matching unambiguous. A projected good pattern must simply be densely textured, photometrically consistent, and have no repetition along the axis of matching, in matching range. To create a favorable pattern for a specific optical configuration and matching algorithm, one can perform optimization over a synthetic pipeline that models a projector design. By modeling both the optical system's physical constraints and realistic imaging noise, one can obtain better texture projectors~\cite{konolige2010projected}. 

\section{RealSense R200 Family}
\begin{figure*}
\begin{center}
\fbox{
	\includegraphics[width=0.3\linewidth]{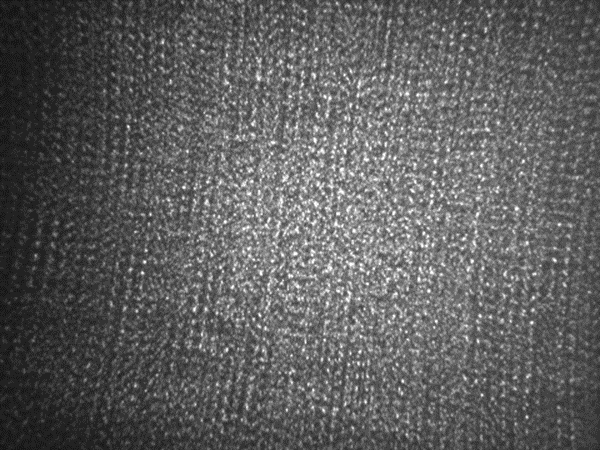}
     \includegraphics[width=0.28\linewidth]{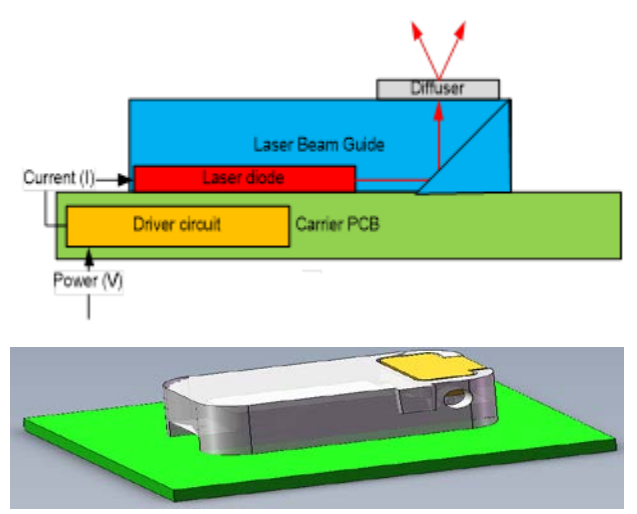}
     \raisebox{0.45\height}{\includegraphics[width=0.3\linewidth]{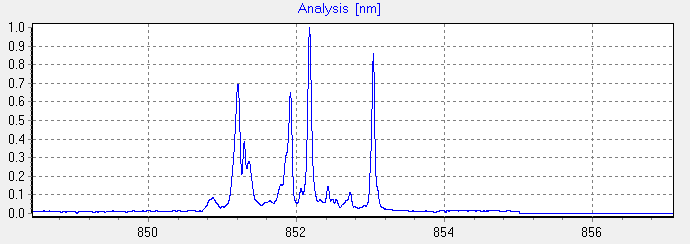}}
}
\end{center}
   \caption{An overview of the R200 pattern projector. From left to right, the spatial structure of the projected pattern, a mechanical overview of the projector design, and an example emission spectrum from a measured unit.}
\label{fig:projector}
\end{figure*}
There have been multiple stereoscopic depth cameras released by Intel. However, many of them share similar (or identical) imagers, projectors and imaging processor. We will refer to all these products as the Intel R200, although this analysis also applies to the LR200 and ZR300.

These modules are connected and powered by a single USB connector, are approximately 100x10x4mm in size, and support a range of resolutions and frame-rates. Each unit is individually calibrated in the factory  to a subpixel accurate camera model for all three lenses on the board. Undistortion and rectification is done in hardware for the left-right pair of imagers, and is done on the host for the color camera.

\subsection{R200 Processor}
The R200 includes an imaging processor, sometimes referred to as the DS4, which follows a long lineage of hardware accelerated stereo correlation engines, from FPGA systems~\cite{DBLP:conf/fccm/WoodfillH97} to  ASICs~\cite{DBLP:conf/cvpr/WoodfillGB04}. These systems are scalable in terms of both resolution, frame-rate and accuracy, allowing variation in optical configuration to result in trade-offs while using only a single processing block~\cite{DBLP:conf/cvpr/WoodfillGJBB06}. 

The image processor on the R200 has a fixed disparity search range (64 disparities), hardware rectification for the left-right stereo pair, and up to 5 bits of subpixel precision. Firmware controls auto-exposure, USB control logic, and properties of the stereo correlation engine. In practice, the processor can evaluate over 1.5 billion disparity candidates per second in less than half a watt of power consumption. In fact, the entire depth pipeline, including the rectification, depth computation, imagers and an active stereo projector, can be run with about one watt of power draw. Following previous designs~\cite{DBLP:conf/cvpr/WoodfillGB04}, the processor is causal and has minimal latency between data input and depth output; on the order of tens of scan-lines.  

\subsection{R200 Algorithm} \label{sec:alg}
Building on top of previous hardware correlation engines \cite{DBLP:conf/cvpr/WoodfillGB04}, the R200 uses a  Census cost function to compare left and right images ~\cite{DBLP:conf/eccv/ZabihW94}. Thorough comparisons of photometric correlation methods showed the Census descriptor to be among the most robust to handling noisy imagining environments~\cite{hirschmuller2009evaluation}. After performing a local 7x7 Census transform, a 64 disparity search is performed, with a 7x7 box filter to aggregate costs. The best-fit candidate is selected, a subpixel refinement step is performed, and a set of filters are applied to filter out bad matches (see section \ref{sec:preset} for details). However, all these filters do is mask out bad candidates. No post-processing, smoothing or filtering is otherwise performed. That is, no matches or depth values are ever ``imagined'' or interpolated. All depth points generated by the chip are real, high-quality photometric matches between the left-right stereo pairs. This allows the algorithm to scale well to noisy images, as shown in Section~\ref{sec:noise}. 

The algorithm is a fixed function block inside the processor and most of its matching properties are not configurable. The settings exposed to users are limited to selecting the input resolution of matching (which determines accuracy, minimum distance, etc.) and configuring the \textit{interest operators} that discard data as invalid (detailed in Section~\ref{sec:preset}).

Compared to many other depth generation algorithms, the R200 is a straightforward stereo correlation engine. It computes best-fit matches between left-right images to the highest accuracy it is able to, and filters out results it has low confidence in. There is no smoothing, clipping, or spatial filtering of data outside of cost aggregation. This avoids artifacts and over-smoothing of data, providing generally higher quality matches than other methods. These comparisons are shown in Section~\ref{sec:algperf}. 

\subsection{R200 Imaging Hardware}
A thorough overview of the R200 hardware is provided in Intel's datasheet~\cite{intel:r200mouser}. We summarize and elaborate on certain performance-relevant details here.  Each unit is individually calibrated in the factory for subpixel accurate rectification between all three imagers. There is a high quality stiffener, along with detailed guidelines on tolerances and requirements to ensure that these cameras hold calibration over their lifetime~\cite{intel:r200mouser}.

\subsubsection{Cameras}
The R200 module includes three cameras: a left-right stereo pair and an additional color camera. The left-right cameras are identical, placed nominally 70mm apart, and feature 10bit, global shutter VGA (640x480) CMOS monochrome cameras with a visible cut filter to block light  approximately below 840nm. This is because the active texture projector is in the infrared band and the imaging system seeks to maximize the SNR of the active texture projector. Thus, without a strong infrared light source such as the projector, incandescent light-bulbs, or the sun, the left-right cameras are unable to see. On the other hand, because of the combination of texture projector and being able to capture sunlight, the R200 is able to work both in 0 lux environments and in broad daylight. The field of view of these cameras is nominally  60 x 45 x 70 (degrees; horizontal x vertical x diagonal). The left-right cameras are capable of running at 30, 60 or 90 Hz.

The color camera on the R200 is a Full HD (1920x1080), Bayer-patterned, rolling shutter CMOS imager. Available on the R200 module is an ISP which performs demosiacing, color correction, white balance, etc.  This camera stream is time-synchronized with the left-right camera pair at start-of-frame. Undistortion of the color camera, unlike the left-right pair, is performed in software on the host. The nominal field of view is  70 x 43 x 77 (degrees; horizontal x vertical x diagonal), and the camera runs at 30 Hz at Full HD, and at higher frame rates for its lower resolutions.

Currently, the R200 family of cameras uses a rectilinear, modified Brown-Conrady~\cite{brown1966decentering} model for all cameras. There are three even-powered radial terms and two tangential distortion terms. The modification comes in the form of computing tangential terms with radially-corrected $x,y$. The open-source library for accessing the Intel RealSense cameras details the distortion model exactly~\cite{librealsense}.The rectification model places the epipoles at infinity, so the search for correspondence can be restricted to a single scan line~\cite{DBLP:conf/cvpr/WoodfillGB04}.

\subsubsection{Projector} \label{sec:proj}

Each R200 also includes an infrared texture projector with a fixed pattern. The pattern can be seen in \figurename~\ref{fig:projector}. The pattern itself is designed to be a high-contrast, random dot pattern. Patterns may vary from unit-to-unit and may slightly change during usage of an individual camera module. Since the projector is laser-based (Class 1), the pattern also exhibits laser speckle. Unfortunately, the laser speckle pattern seen from the left and right imagers is unique to each view; this creates photometric inconsistency. This inconsistency can cause artifacts in the matching, either creating matches where there are none or adding correlation noise to high-quality matches. Thus the R200 includes several techniques to mitigate laser speckle artifacts, some of which are seen in \figurename~\ref{fig:projector}, featuring wavelength and bandwidth diversity.

\section{Performance} 
We decouple our discussion of performance into algorithmic performance expectations and system performance expectations. The algorithmic portion quantifies the algorithm design and its performance on standard stereo datasets; given that this is a local algorithm with block correlation \cite{hirschmuller2009evaluation}, its near state-of-the-art performance on some metrics may be surprising. The system performance section focuses on how actual R200 family units perform on real world targets, which we hope is useful for those building algorithms that consume data from these camera systems. 

\subsection{Algorithm Performance}\label{sec:algperf}
In the design of stereo correspondence algorithms, there exist multiple standard datasets in use. The two most common are the KITTI ~\cite{Geiger2012CVPR}, and Middlebury datasets ~\cite{scharstein2014high}. In our testing and development we prefer the Middlebury dataset for multiple reasons. Middlebury is a higher resolution dataset, with subpixel accurate disparity annotations (KITTI tends to use 3 disparities as its threshold), and dense annotations for both data and occluded regions (KITTI data comes from a sparse LIDAR scan). All of these properties make the Middlebury dataset better suited to evaluation for a metrically accurate RGBD sensor, where both density and subpixel precision (as seen in Table~\ref{tab:perf}) are a key part of performance.

\subsubsection{Middlebury Results}
The Middlebury dataset  includes multiple metrics, and of these, we focus on what Middlebury calls \textit{sparse} results, where an algorithm is allowed to discard data it feels it is unable to compute correctly. This is for multiple reasons, including the fact that the R200 laser projector disambiguates  challenging cases, but primarily because the ability to remove outliers and mismatches is in fact, a key component of measuring algorithm quality. RGBD camera users do not simply want the highest quality correlations, but also removal of noisy or incorrect data.

\begin{table}
\begin{center}
\begin{tabular}{|l|c|c|c|c|c|}
\hline
Metric  & bad0.5 & bad1.0 & bad2.0 & bad4.0 & A50 \\
  \hline\hline
Ranking & 2nd & 3rd & 3rd & 4th & 1st\\
\hline
\end{tabular}
\end{center}
\caption{R200 algorithm ranking on the on Middlebury training sparse dataset, when configured with the high-quality preset. There are presently 61 algorithms listed.}
\label{tab:rank}
\end{table}

When targeting high quality results (using the high preset described in Section~\ref{sec:preset}), a summary of the R200's performance on the Middlebury images is described in Table~\ref{tab:rank}. When looking at how well the R200 algorithm is able to generate accurate results, we can see that the R200 is able to provide near state-of-the-art performance. For example, using the median disparity error metric (Table~\ref{tab:a50}), the R200 is the best performing algorithm, with a 15\% improvement compared to the second best algorithm, although its data density is lower than other top-ranked results. On the other hand, when looking at the fraction of pixels with less than half a pixel of disparity error (Table~\ref{tab:bad5}), the R200 algorithm would be second-best in the Middlebury rankings, despite having higher density than the third-ranked result. 

These results are somewhat surprising, as the R200 is local matching algorithm implemented on an ASIC, with only a few scan-lines of latency, whereas other comparison algorithms use full-frame, global and semi-global techniques to generate robust matches. This suggests that the R200 features carefully designed interest operators and subpixel methods, as described in Section~\ref{sec:alg}. 

\begin{table}
\begin{center}
\begin{tabular}{|l|c|c|}
\hline
Algorithm  & bad0.5 & Validity \\
  \hline\hline
SED~\cite{Pena2017} & 0.87\% & 0.9\% \\
R200 &  8.39\%  & 17.6\% \\
ICSG~\cite{shahbazi2016revisiting} & 10.32\% & 16.5\% \\
\hline
\end{tabular}
\end{center}
\caption{Top ranked Middlebury training sparse results for the bad 0.5 metric, the fraction of valid disparities that are more than 0.5 disparities incorrect.}
\label{tab:bad5}
\end{table}

\begin{table}
\begin{center}
\begin{tabular}{|l|c|c|}
\hline
Algorithm  & A50 & Validity \\
  \hline\hline
R200 &  0.25 px & 17.6\% \\
R-NCC & 0.30 px & 31.4\% \\
3DMST~\cite{mst2017} & 0.30 px & 100\% \\
MCCNN\_Layout & 0.31 px & 100\% \\
\hline
\end{tabular}
\end{center}
\caption{Top ranked Middlebury training sparse results for the A50 metric (median disparity error).}
\label{tab:a50}
\end{table}

\subsubsection{Noise Resilience} \label{sec:noise}
\begin{table}
\begin{center}
\begin{tabular}{|l|c|c|c|c|}
\hline
Algorithm  & Dataset & bad0.5 & avgErr  & Validity \\
  \hline\hline
ELAS~\cite{geiger2010efficient} & MQ  &  25.0\%  & 3.5 px & 78\% \\
ELAS~\cite{geiger2010efficient} & MQN &  33.7\%  & 11.5 px & 63\% \\
  \hline\hline
R200 & MQ  &  12.9\%  & 4.9 px & 68\% \\
R200 & MQN &  13.1\%  & 5.6 px & 60\% \\
\hline
\end{tabular}
\end{center}
\caption{Comparing results between the standard Middlebury data (MQ)  against a noisy version of the data (MQN). These are on quarter sized  Jadeplant image. While ELAS degrades severely, the R200 algorithm is nearly robust to this additional level of noise. }
\label{tab:noise}
\end{table}
There exists a very large gap in the quality of images available in standard stereoscopic datasets such as Middlebury, and the portable, compact image sensors used on the R200. Middlebury images are collected with a high-quality DSLR camera with large pixels and good optics. Whereas the R200 module features compact, low-cost, webcam-quality CMOS cameras. The R200 stereoscopic matching algorithm is optimized to work better on such noisy image data, but there isn't a standard evaluation process for this type of data.

Thus we construct our own Middlebury Noisy dataset, which simply adds realistic imaging noise~\cite{hasinoff2010noise}, in the form of photon noise, read noise and a Gaussian blur to model MTF differences. The magnitude of added noise roughly corresponds to the noise level of the R200 CMOS image sensors. The comparison of how well the R200 algorithm performs on both the standard and noisy Middlebury Quarter-sized images is shown in Table~\ref{tab:noise}. The results show that the R200 algorithm is only minimally affected, especially in comparison to the reference ELAS~\cite{geiger2010efficient} algorithm provided by the Middlebury evaluation. 

\subsection{System Performance} \label{sec:perf}
\begin{table}
\begin{center}
\begin{tabular}{|l|c|c|}
\hline
Metric  & Symbol & Value \\
  \hline\hline
1\% RMS Range & $r_{\epsilon_{\text{\% of z}} = 1}$ & 4.1 m \\
95\% Fill Range & $r_{\rho = 95\%}$ & 6.0 m \\
Best Case RMS & $\epsilon_{\text{mm}}$ & 1.0 mm \\
RMS at 1 meter & $\epsilon_{\text{mm}}$ & 2.1 mm \\
RMS at 2 meter & $\epsilon_{\text{mm}}$ & 8.0 mm \\
Absolute Error at 1 m & $\epsilon_{\text{calib}}$ & 5 mm \\
X-Y Detectable Size & $\epsilon_{x y}$ &     \begin{tabular}{@{}c@{}}5 pixels \\ 1\% of distance \end{tabular} \\
Minimum Distance & $z_{\min}$ & 0.53 m \\
Dynamic Range & $\frac{I_{\text{max}}}{I_{\text{min}}}$ & 40x \\
Disparity RMS (static) & $\epsilon_{\text{d}}$ & 0.08 pixel \\
Disparity RMS (moving) & $\epsilon_{\text{d}}$ & 0.05 pixel\\
\hline
\end{tabular}
\end{center}
\caption{Measured performance of an example R200 unit at 480x360 resolution at 30 Hz. Definitions are elaborated in section \ref{sec:perf}.  }
\label{tab:perf}
\end{table}
We use white walls to test our unstructured light systems, and textured walls to measure passive systems. Due to the issues explained in section \ref{sec:motion}, the R200 obtains higher accuracy results in passive, well-illuminated conditions. A summary of results is available in Table~\ref{tab:perf}. To avoid over-sampling due to distortion loss, most of our measurements are done at 480x360 resolution instead of the native 640x480 resolution. 
\begin{figure}[b]
\begin{center}
\fbox{
	\includegraphics[width=0.45\linewidth]{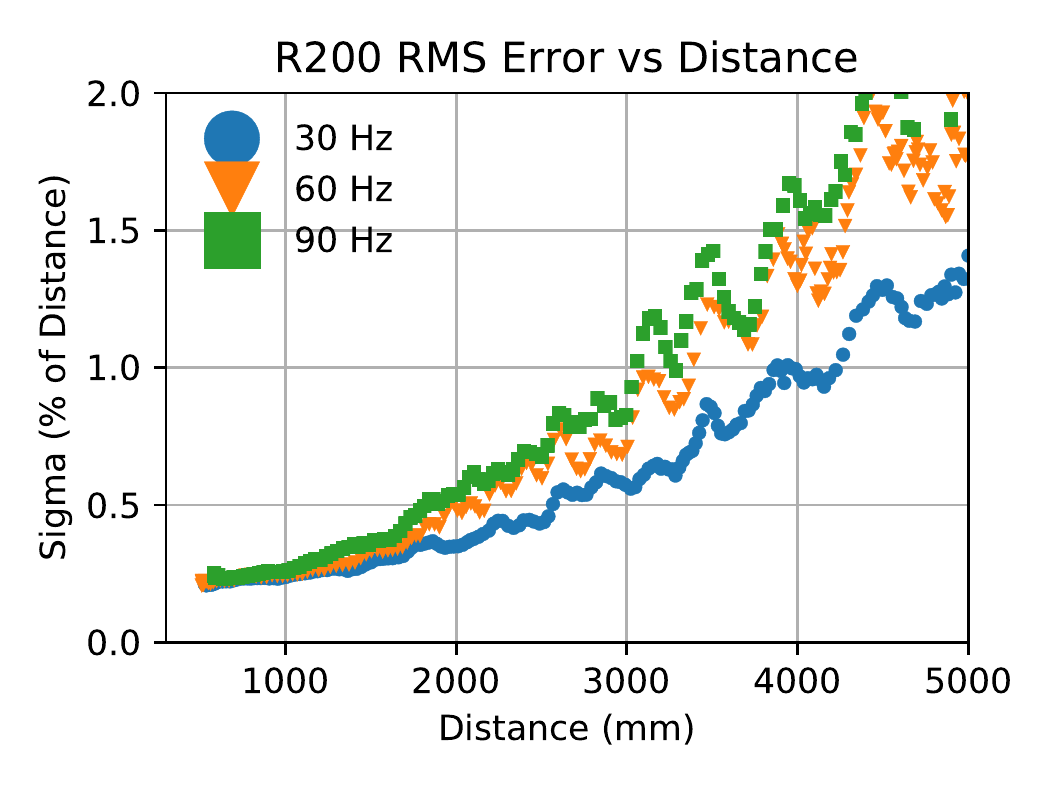}
     \includegraphics[width=0.45\linewidth]{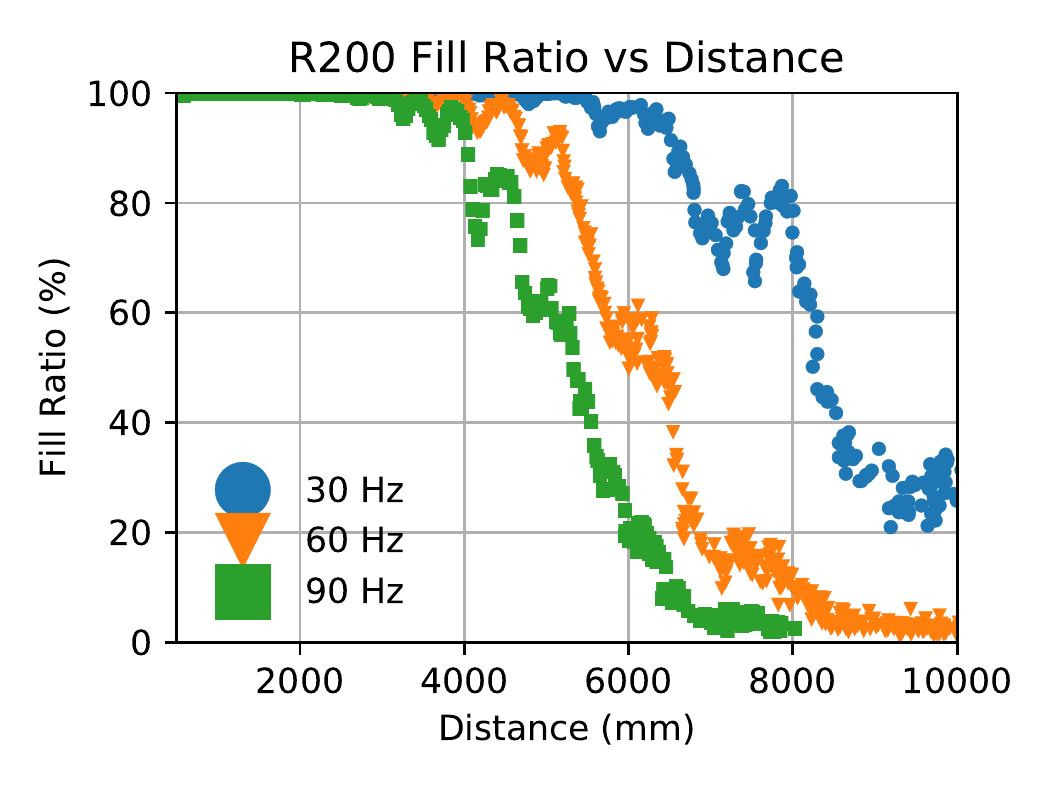}
}

\end{center}
   \caption{R200 performance at different frame-rates, across different distances. On the left is RMS error from a best-fit planar surface, on the right is the density of measurements.}
\label{fig:rate}
\end{figure}
\subsubsection{Depth Range}
The minimum distance that the R200 can detect is a function of its fixed disparity search range. Due to the relationship in equation~\ref{eq:dispdepth}, this is a hard limit in real-world space, of around $\frac{1}{2}$ a meter at 480x360. At lower resolutions, the minimum distance moves closer to the camera as the fixed range covers a larger fraction of the field-of-view. For example, it is roughly $\frac{1}{3}$ of a meter at 320x240. 

Since the R200 is a stereoscopic matching system, its maximum distance varies depending on illumination and texture conditions. In passive conditions, for a well textured target, $d=1$ corresponds to roughly 30 meters at 480x360, and in a strict sampling sense, this is the last non-infinity distance disparity, and is one definition of maximum z. In indoor conditions, with the R200 projector being the only source of texture, SNR tends to be limiting factor in determining distance. The metric we use in this paper is greatest distance at which a perpendicular white wall returns greater than 95\% of its measurements in the center of the FOV. This distance is roughly 6 meters at 30 Hz. See Section~\ref{sec:extra} for how this measurement can be used to estimate performance under other measurement conditions.

Higher frame-rates lead to shorter exposure times and hence a decrease in SNR-limited range. This scales with roughly with the square-root of frame-time in photon-limited environments, as can be seen with 30, 60 and 90 Hz measurements in \figurename~\ref{fig:rate}. 

\subsubsection{RMS Error} \label{sec:rms}
\begin{figure}
\begin{center}
\fbox{
	\includegraphics[width=0.3\linewidth]{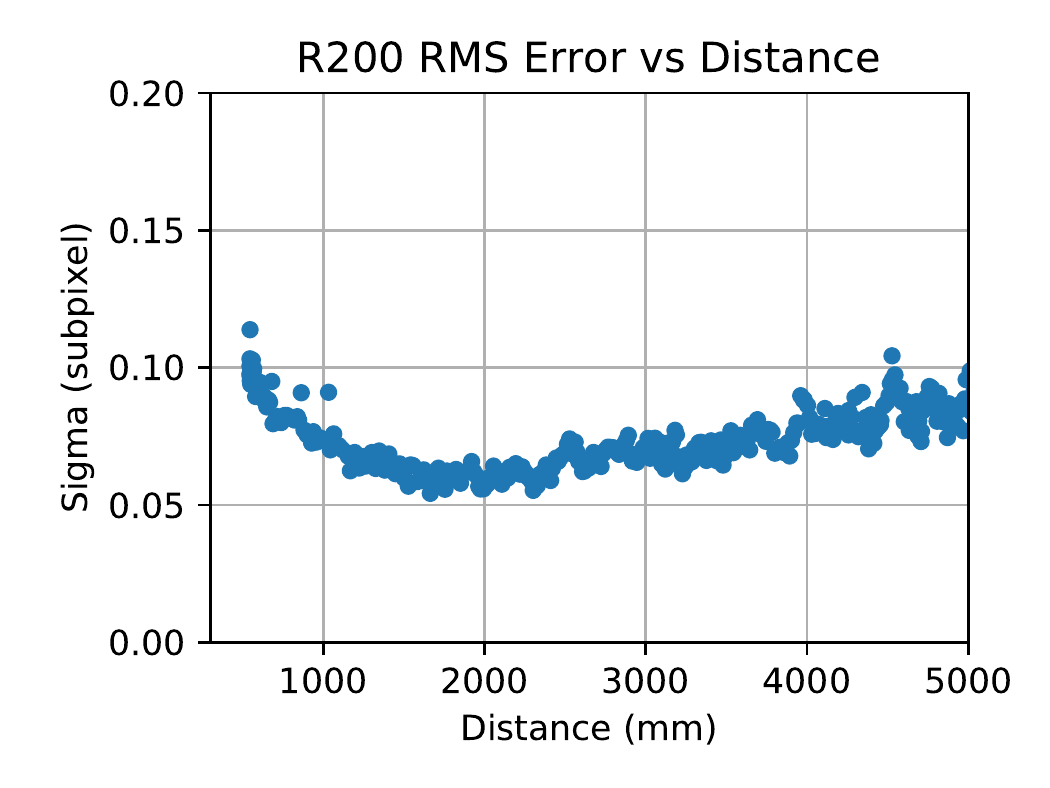}
     \includegraphics[width=0.3\linewidth]{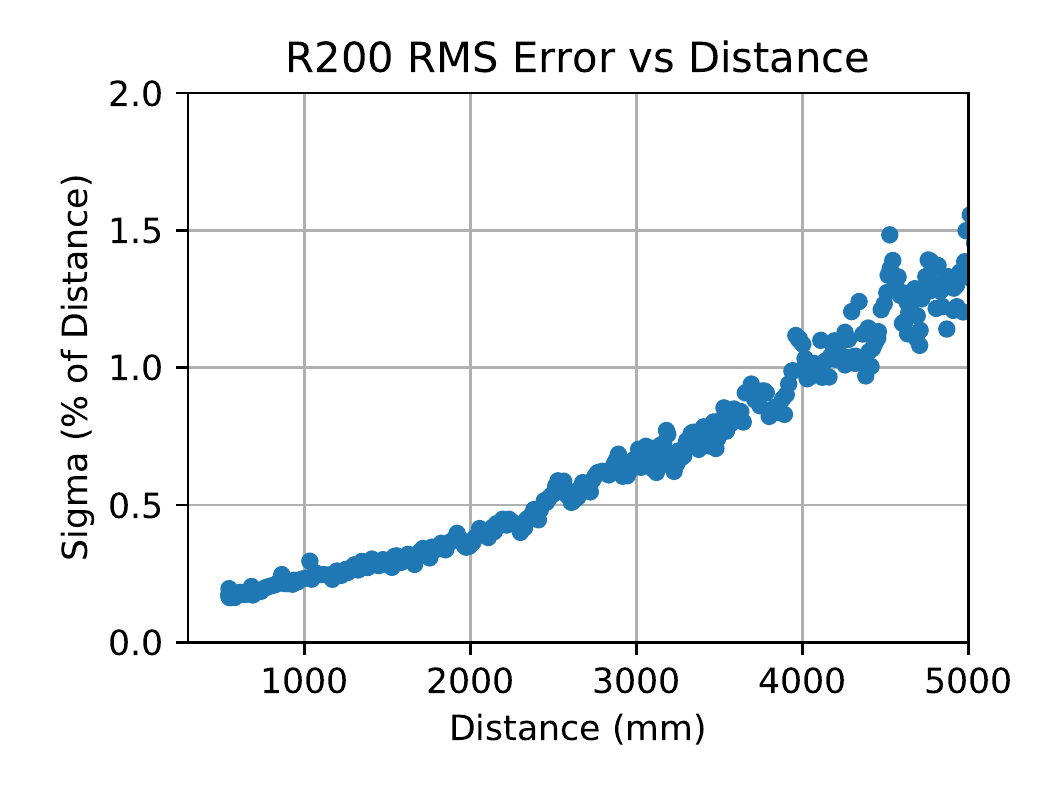}
     \includegraphics[width=0.3\linewidth]{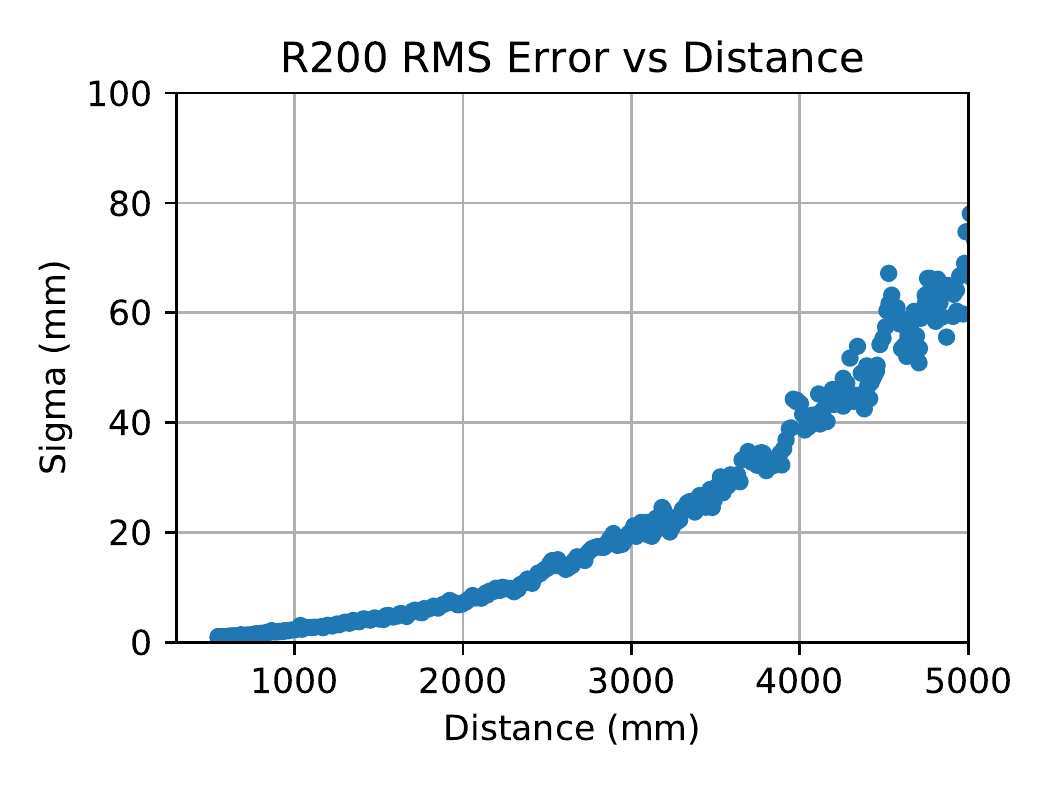}
}

\end{center}
   \caption{RMS depth noise over distance is roughly constant in disparity space ($\epsilon_{\text{pixel}}$), linear as a fraction of distance ($\epsilon_{\text{\% of z}}$), and quadratic in real-world units ($\epsilon_{\text{mm}}$) }
\label{fig:acc}
\end{figure}

To estimate the noise and accuracy of an R200, we again use the wall targets described earlier. A best-fit plane is fit to the data and the RMS error from the plane is computed \footnote{This is done with an SVD in 3D Cartesian space, and the smallest singular value corresponds to RMS error from the best-fit plane. }. These measurements are shown in \figurename~\ref{fig:acc} and Table~\ref{tab:perf}, and are SNR limited even at 30 Hz, where faster frame-rates show more error, see \figurename~\ref{fig:rate}. Of note is that the R200 RMS noise is below that of pixel ($\approx 0.1$ disparities), in line with previous measurements of the R200 accuracy~\cite{ryan2016hyperdepth}. These results highlight the importance of using a high-quality subpixel interpolation method. Traditional curve fitting approaches have failure cases that sophisticated methods are capable of addressing~\cite{nehab2005improved}.

Unit conversions can make these RMS numbers easier to work with. Simple manipulations of equation~\ref{eq:derr} give the following convenient expressions for depth accuracy as a function of distance, corresponding to the results in \figurename~\ref{fig:acc},
\begin{equation} \label{eq:convert}
\epsilon_{\text{\% of z}} = \frac{\epsilon_{mm}}{z}  = \epsilon_{\text{d}} \frac{z}{f B}.
\end{equation}
\subsubsection{Spatial Accuracy}
The spatial accuracy (the accuracy of edges in images space) is often a weakness of stereoscopic matching systems based their need for spatial support structures. In the R200's case, with a fixed block size, this behavior is predictable and symmetric in $x$ and $y$, with the R200 incorrectly merging gaps between edges of less than 5 pixels, or, in world units, about 1\% of distance at 480x360 resolution. 

\subsubsection{Calibration Tolerances}
Due to imperfections in fitting an undistortion and rectification model to the cameras, the R200 camera may consistently return depth that has an offset relative to the real world. This may be scale or distance dependent depending on which part of the model has fitting noise. Our measurements of several R200s show that this calibration error is zero mean, and typically small, on the order of 5mm at 1 meter. Additionally, the R200 is subject to stereo bias \cite{freundlich2015exact} but a planar derivation of that bias can be calculated as $\epsilon = -\frac{d}{2 d^2-1}$ (for $d > \sqrt[]{2}$), and this is dominated by imaging noise for ranges below 10 meters on the R200. 

\subsubsection{Dynamic Range}
By sweeping a camera exposure from its fastest exposure and smallest gain to its longest exposure and longest gain, we can measure at what points we are able to get 95\% density. This ratio, from the darkest condition where data is returned accurately (due to SNR bounds), to the brightest condition where it is returned accurately (due to saturation), is the system's dynamic range. This is dependent on the choice of emitter, imagers, optics, and algorithm. Our measurements show this factor to be roughly 40x on most R200 units.

\subsubsection{Reducing R200 noise} \label{sec:motion}
Once an R200 is placed at an appropriate distance from a properly illuminated target, there are still two primary sources of data noise that limit a user's ability to acquire high quality depth, laser speckle and data-dependent noise. Both of them can be mitigated with physical motion. 

\begin{figure}
\begin{center}
\fbox{
	\includegraphics[width=0.3\linewidth]{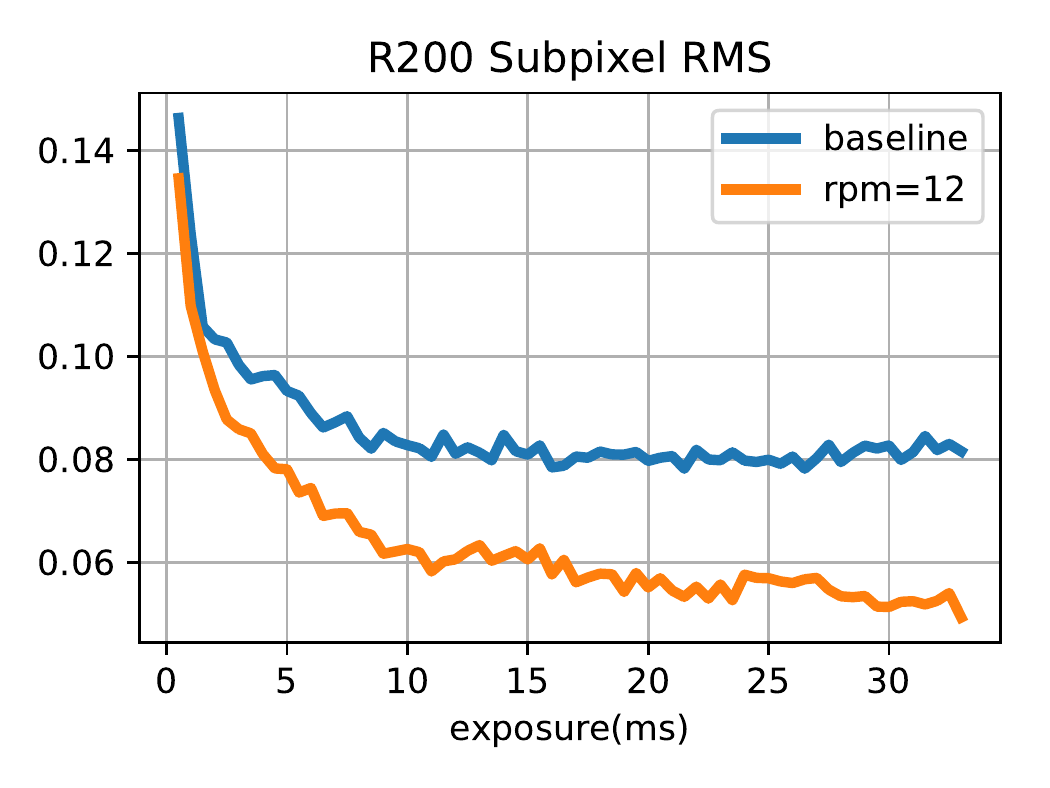}
     \includegraphics[width=0.3\linewidth]{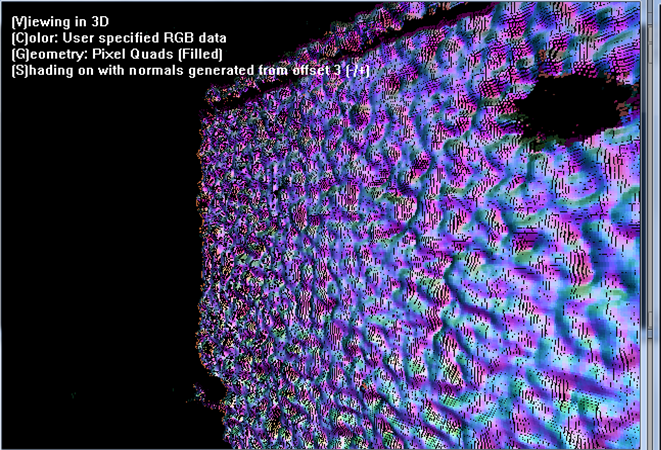}
     \includegraphics[width=0.3\linewidth]{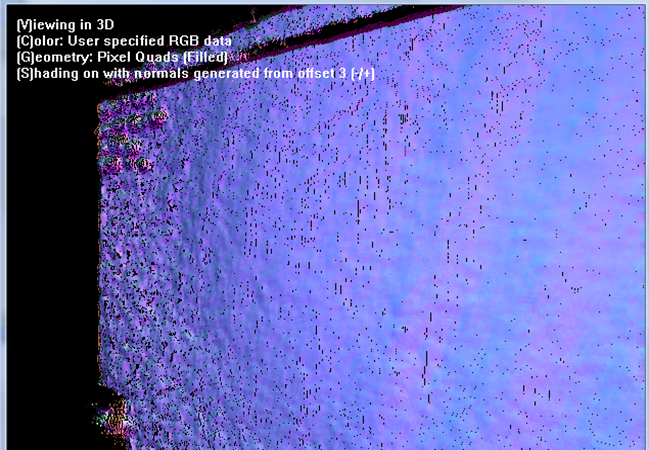}
}

\end{center}
   \caption{The left-most image is a graph of measured $\epsilon_d$ versus exposure time for a spinning vs static target. The right pair of images are the results of temporal averaging with static camera and with a moving projector. }
\label{fig:motion}
\end{figure}

Subjective speckle, as outlined in Section~\ref{sec:proj}, contributes noise into the matching algorithm and limits the SNR of capturing the projected pattern. However, laser speckle will start to disappear if either the camera or the target is in motion. When the camera moves relative to the scene, RMS error decreases and data density increases. When doing the experiments in \figurename~\ref{fig:rate}, the camera was hand-held and hence in motion; if this experiment is repeated on a tripod, the results will be worse. This effect can be isolated by comparing the RMS error (over a sweep of exposure) of both a static wall target and one with a spinning target. These result are shown in \figurename~\ref{fig:motion}, and we can see a nearly 40\% reduction in RMS error at 33 millisecond exposure time when the target is spinning at 12 rpm.

When looked at closely, the R200 tends to have ``bumpy'' looking data, such as that seen in the center of \figurename~\ref{fig:motion}. This is due to subpixel matching noise, and the high level of subpixel precision (5 bits) returned by the R200. To generate the \textit{plating} behavior seen on other subpixel estimating RGBD sensors such as the ASUS Xtion, one can simply clip the R200 data above its baseline noise level, at a lower subpixel precision (such as 2 bits). A further problem is that the image in the center of \figurename~\ref{fig:motion} is actually the temporal average of a static scene and static camera (over 1,000 frames). This fixed pattern noise stems from having fixed input images, and hence fixed subpixel artifacts. In order to obtain the results seen on the right in \figurename~\ref{fig:motion}, one needs to simply move the projector while performing the time-averaging, and then the R200 returns zero mean noise for every pixel over time. An equivalent operation would be to move the camera relative to a scene, while performing registration and aggregation~\cite{niessner2013hashing}. 

Additionally, the ripples seen in \figurename~\ref{fig:rate} correspond to increases and decreases in performance as the camera moves from sampling integer disparity locations (better performance) to half pixel disparity shifts between the left and right images (worse performance). The dramatic effect on density comes from an aggressive preset configuration.

\subsubsection{Presets} \label{sec:preset}
Almost all depth systems natively generate a depth for every pixel. It's the responsibility of the algorithm to find cases where it has low-quality matches. However, as these are noisy measures, they generally require trading off sensitivity and specificity. This can be seen as walking along the receiver operating characteristic. The R200 is able to support this mode of operation by controlling how aggressive the algorithm is at discarding matches. 

To measure the quality of a match, a set of \textit{interest operators}~\cite{DBLP:conf/cvpr/WoodfillGB04} is implemented, similar to those used in standard  matching literature~\cite{hu2012quantitative}.  They are, a minimum matching score, maximum matching score, left-right threshold (in $\frac{1}{32}$ pixels), neighbor threshold, second-peak threshold, texture threshold (at least $N$ pixels with difference $\delta$ in a local window), and median threshold (technically a percentile estimator, computed stochastically \cite{robbins1951stochastic}). 

Each of these measures has a threshold and if pixel fails any of them, it is marked invalid. There is no confidence map provided, data is either returned or marked zero to indicate that it is invalid. Raw costs are poorly correlated with matching quality, and aggregation of multiple features into combined threshold is a classification problem \cite{haeusler2013ensemble,motten2011binary}.

\begin{figure}
\begin{center}
\fbox{
	\includegraphics[width=0.3\linewidth]{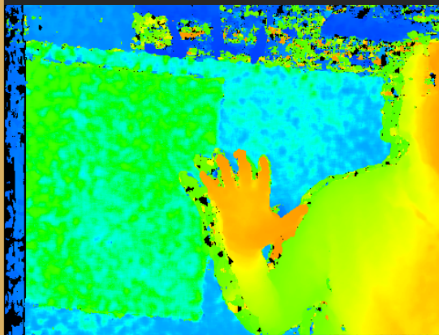}
     \includegraphics[width=0.3\linewidth]{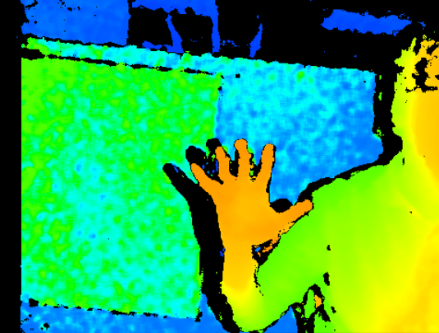}
     \includegraphics[width=0.3\linewidth]{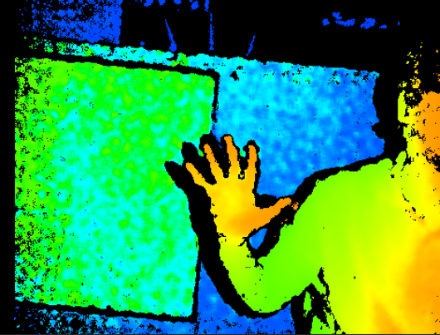}
}

\end{center}
   \caption{A visual example of the Off, Medium and High presets on an R200}
\label{fig:presets}
\end{figure}

To enable users the ability to trade-off density for accuracy, the R200 software provides presets, or optimized threshold sets, which constitute an example Pareto set of configurations. Each of these presets is a (nearly) Pareto optimal configuration over a training dataset for a given trade-off of accuracy and density when using R200 image data. Using these presets, one can get nearly a 100\% dense image, with many erroneous matches, or a very sparse image with very few bad matches as seen in \figurename~\ref{fig:presets}. 

Additionally, this trade-off of accuracy and density can be seen in some quantitative metrics computed using these presets, shown in table ~\ref{tab:presets}. It is clear that \textit{higher} presets, which are more aggressive, have fewer false positives and produce more accurate data, but suffer a decrease in data density (or equivalently, the range at which a density threshold is satisfied). 
\begin{table}
\begin{center}
\begin{tabular}{|l|c|c|c|}
\hline
Preset  & FPR & $r_{\rho = 95\%}$ & $\max(\sigma_x,\sigma_y) > \sigma_z$ \\
  \hline\hline
Off & 91.3\% & 7.0 m & 6.1 m \\
Low & 19.8\% & 6.9 m & 6.6 m\\
Medium & 5.8\% & 5.8 m & 6.7 m \\
High & 0.5\% & 4.2 m & 6.8 m\\
\hline
\end{tabular}
\end{center}
\caption{Performance of different R200 present configurations. FPR, false positive rate, is a percentage of data returned when given a scene below minimum z. $r_{\rho = 95\%}$ and $\max(\sigma_x,\sigma_y) > \sigma_z$ columns contain distances where those data conditions are satisfied, in meters. Further is better.}
\label{tab:presets}
\end{table}

\subsubsection{Extrapolating Performance} \label{sec:extra} 
Based on some of the simple measurements computed in Section~\ref{sec:perf}, which use simply a flat white wall, we are able to extrapolate the expected performance of the R200 into other conditions and domains. This is possible because the 95\% density metric ($r_{\rho = 95\%}$) is indicative of the algorithm being at an SNR boundary. With indoor conditions that have little ambient infrared, we can then use basic physics to estimate how well the camera will perform under different measurement conditions. 
$$r_{\text{expected}} \approx r_{\rho = 95\%} \sqrt{\cos(\theta_{\text{target}}) \cdot \alpha \cdot \cos(\theta_{\text{FOV}})^7}$$
$\alpha$ is albedo or reflectance. $r_{\rho = 95\%}$ is how far you can image a white wall with 95\% density (meters). $\theta_{\text{FOV}}$ is how far off camera axis you are. $\theta_{\text{target}}$ is the tilt of your target relative to the camera plane, following Lambertian reflectance. The seventh power comes from a $\cos(\theta)^4$ lens vignetting and approximately a $\cos(\theta)^3$ projector falloff. The square root comes from the inverse square law. For example, if we have a dark ($\alpha = 0.2$) floor, in the lower quarter of the vertical FOV ($\theta_{\text{FOV}} = 15^\circ $), and is tilted relative to the camera ($\theta_{\text{target}} = 60^\circ $), we'd get $r_{\text{expected}} \approx 1.4 \text{m}$. This matches measured performance, where dark, tilted floors in the corners of the field-of-view are not visible at long distances in the R200 depth stream. Performance issues such as this are greatly mitigated in the D400, as explained in Section~\ref{sec:ds5}. 

\subsubsection{Post-Processing} \label{sec:postproc}
In developing the software shown in Section~\ref{sec:apps}, there have been several properties of the R200 RGBD data we have found useful. First, many approaches struggle with the sparse noise that occurs with the R200's local stereo algorithm; median filters or speckle removal filters (small connected component tests) are efficient at removing it.

Additionally, the use of the color data to smooth and inpaint the depth data with an edge-preserving blur is often helpful, such as with a domain transform \cite{gastal2011domain}. Such filters are best applied in disparity space, not depth space, as the nonlinear relationship between them leads to different results with most  methods. As shown in Section~\ref{sec:rms}, errors are roughly constant in disparity space (with RMS equal to $\epsilon_{d}$), enabling more computationally and theoretically tractable denoising when operating in disparity space.

Lastly, as seen in \figurename~\ref{fig:motion}, when the R200 is at the limits of its matching accuracy, the data exhibits a bumpy characteristic. These bumps have standard deviation equal to $\epsilon_{d} = 0.08$. Due to scene texture and spatial aggregation, this noise is temporally-consistent and spatially smooth, respectively. To obtain smooth surfaces instead, one can simply quantize the data in disparity space, to some $\epsilon_{q}$, where $\epsilon_{d} < \epsilon_{q} \leq 1$; this is done in other commercial depth sensors, and we typically set $\epsilon_{q}$ to be 4 times that of $\epsilon_{d}$.



\section{RealSense D400 Family} \label{sec:ds5}
Intel has announced a follow up to the R200 family RGBD sensors, the D400. These follow the same basic principle of portable, low-latency, metrically accurate, hardware-accelerated stereoscopic RGBD cameras. They support the use of unstructured light illumination, as well as additional time-synchronized camera streams. The basic analysis, metrics, artifacts, and properties outlined above hold true for the D400. Its matching algorithm is a configurable superset of the method used on the R200, producing a significant improvement in results. The design focus is similar to that of the R200, working with noisy image data, and emphasizing accurate, false-positive free depth data.

\begin{figure}
\begin{center}
\begin{framed}
	\includegraphics[width=0.49\linewidth]{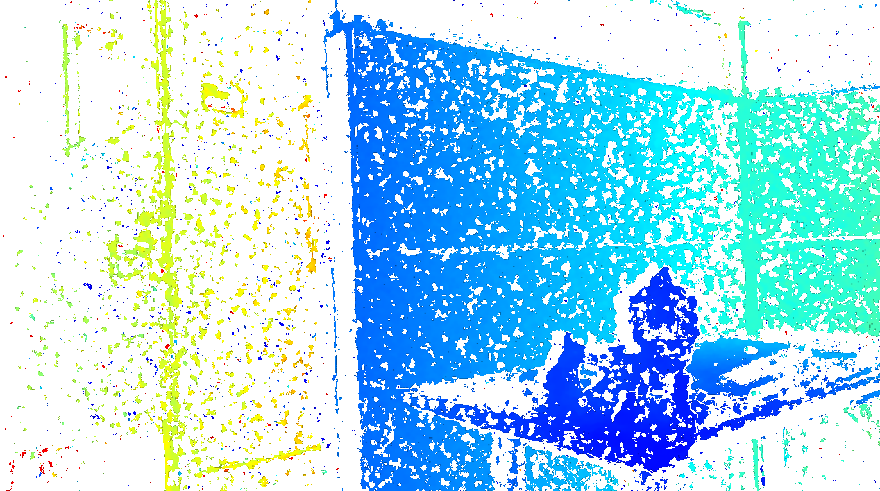}
     \includegraphics[width=0.49\linewidth]{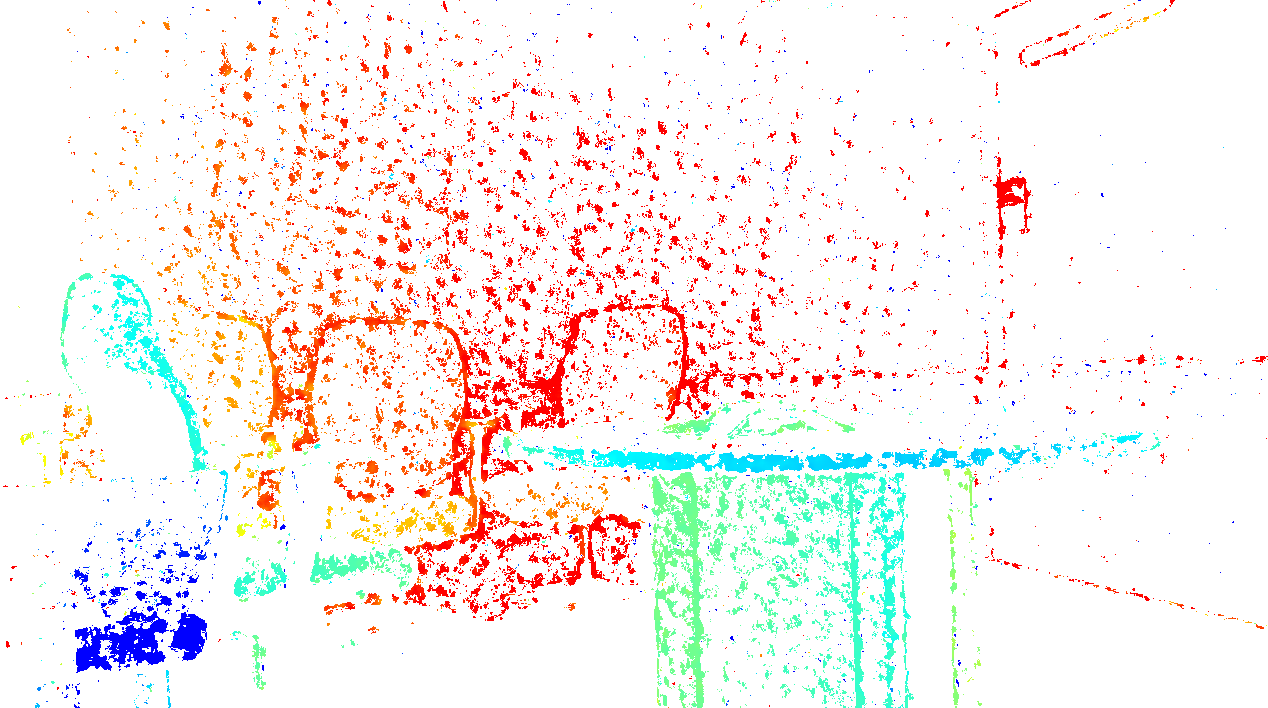} \\
     \vspace{5mm}
	\includegraphics[width=0.49\linewidth]{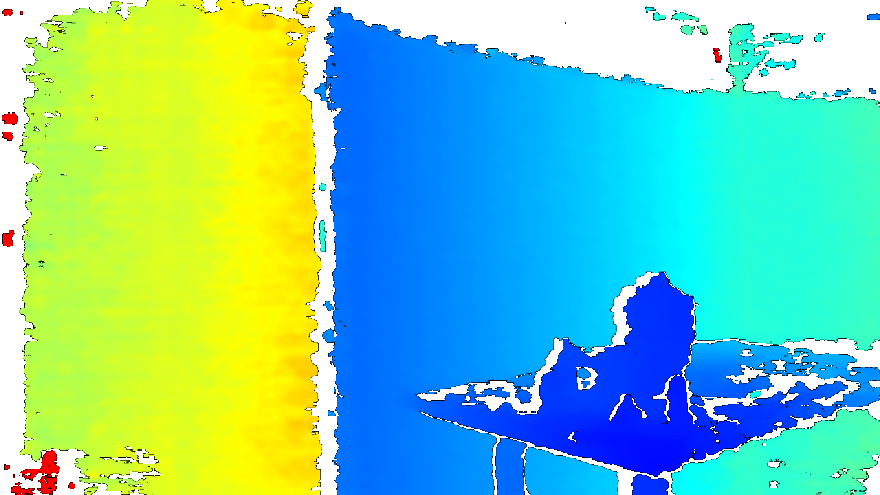}
     \includegraphics[width=0.49\linewidth]{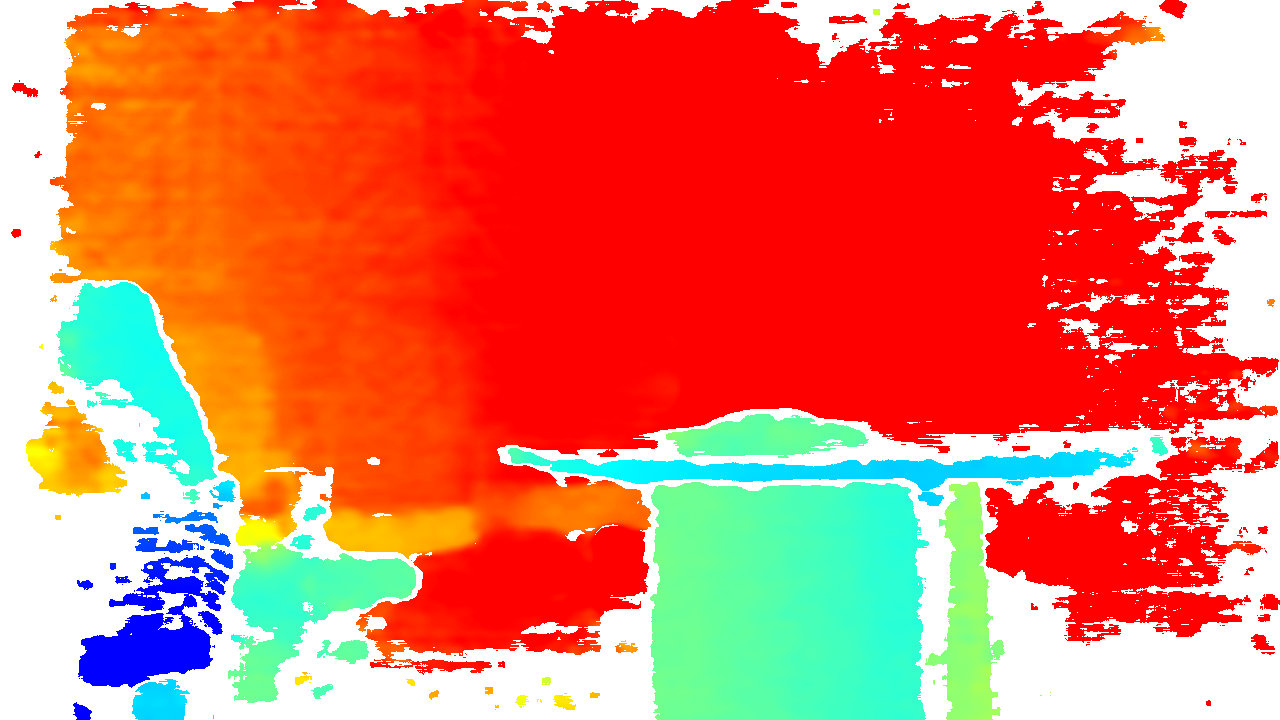}
\end{framed}
\end{center}
   \caption{Top row is R200 algorithm, bottom row is D400 algorithm. Input images are identical. The quality of depth  obtained with D400 is superior to the R200.}
\label{fig:ds5}
\end{figure}

The basic improvements to the D4 Vision Processor include several more recent innovations in stereoscopic matching algorithms. A qualitative example of the improvements these innovations bring is shown in \figurename~\ref{fig:ds5}. The algorithm has been expanded to include various techniques to move past local-only matching and intelligently aggregate a pixel's neighbors estimates into a final estimate. Examples of published techniques that accomplish this are semi-global matching~\cite{hirschmuller2008stereo} and edge-preserving accumulation filters \cite{yang2013full,zhang2009cross}. Additionally, the correlation cost function has been expanded beyond simple Census correlation, integrating other matching measures~\cite{mei2011building}. The D400 features support for larger resolutions, along with a corresponding increase in disparity search range.  Additionally, optimizations in ASIC design allow the D400 family to do all of this while being lower power than the R200, when run on the same input image resolutions. However, at the time of this publication, there are no commercially available D400 units, so we are unable to provide a quantitative analysis of performance like that in Section~\ref{sec:perf}.  


\section{Software and Applications} \label{sec:apps}
\begin{figure}[h]
\begin{center}
\fbox{
   \includegraphics[width=0.5\linewidth]{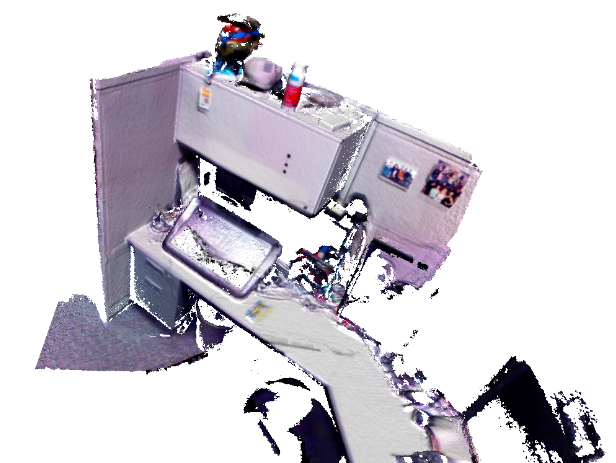} 
   \includegraphics[width=0.4\linewidth]{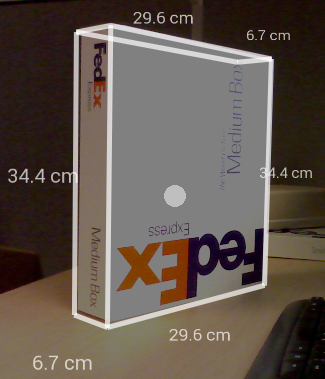}
   }
\end{center}
   \caption{Application examples of the R200 depth cameras. The left image features a volumetric integration algorithm~\cite{rusinkiewicz2002real}, while the right image features a single-frame box dimension estimation algorithm~\cite{youtubebox}.}
\label{fig:app}
\end{figure}

Intel provides a free, proprietary SDK~\cite{intel:r200sdk}, and an open-source, Apache licensed, cross-platform library for streaming camera data~\cite{librealsense}.
A thorough overview of the R200's usage is provided in Intel's SDK presentation ~\cite{intel:r200sdk}. Recently, this R200 family of depth cameras was considered best-in-class for 3D skeletal tracking of hand-pose, praised for its high accuracy and motion tolerance~\cite{Melax:2013:DBS:2532129.2532141}. Additionally, these cameras have been shown to be capable of high-quality 3D volumetric reconstruction~\cite{niessner2013hashing} and measurement of exact object dimensions (see \figurename~\ref{fig:app}). The R200 has also been integrated into several commercially hardware platforms, including the Yuneec Typhoon H multirotor, the ASUS Zenbo home robot, the HP Spectre X2 computer, the TurtleBot3 robotics platform and others. 

\section{Conclusion}
In this work, we have explored the design, properties and performance of the Intel stereoscopic RGBD sensors. We have profiled the stereo matching algorithm performance on reference datasets, along with end-to-end system performance of the R200. In this, we have discussed various performance  challenges in the system, and demonstrated available mitigation strategies. We show how such RGBD sensors work in a variety of situations including outdoors.  

\section{Acknowledgements}
The R200 and D400 depth cameras are the work of Intel's Perceptual Computing Group. We'd like to highlight some of them for their valuable contributions and discussions --- Ehud Pertzov, Gaile Gordon, Ron Buck, Brett Miller, Etienne Grossman, Animesh Mishra, Terry Brown, Jorge Martinez, Bob Caulk, Dave Jurasek, Phil Drain, Paul Winer, Ray Fogg, Aki Takagi, Pierre St. Hilaire, John Sweetser, Tim Allen, Steve Clohset.


\clearpage
{\small
\bibliographystyle{ieee}
\bibliography{egbib}
}

\end{document}